\documentclass[10pt,twocolumn,letterpaper]{article}

\usepackage[pagenumbers]{cvpr} 

\usepackage{graphicx}
\usepackage{amsmath}
\usepackage{amssymb}
\usepackage{booktabs}
\usepackage{algorithm}
\usepackage{algorithmicx}
\usepackage{algpseudocode}
\usepackage{lipsum}
\usepackage{url}

\usepackage{mathtools}

\usepackage{tabularx}
\usepackage{multirow}

\usepackage{trimclip}

\usepackage{graphicx}
\usepackage{booktabs}

\usepackage{listings}
\usepackage{upquote}  
\usepackage{xcolor,colortbl}

\usepackage{listings}
\usepackage{upquote}  
\usepackage{xcolor,colortbl}
\lstdefinestyle{overleaf}{
    backgroundcolor=\color[rgb]{0.95,0.95,0.92},   
    commentstyle=\color[rgb]{0,0.6,0},
    keywordstyle=\color{magenta},
    numberstyle=\tiny\color[rgb]{0.5,0.5,0.5},
    stringstyle=\color[rgb]{0.58,0,0.82},
    basicstyle=\ttfamily\footnotesize,
    breakatwhitespace=false,         
    breaklines=true,                 
    captionpos=b,                    
    keepspaces=true,                 
    numbers=left,                    
    numbersep=5pt,                  
    showspaces=false,                
    showstringspaces=false,
    showtabs=false,                  
    tabsize=2
}
\lstdefinestyle{mocov3}{
  backgroundcolor=\color{white},
  basicstyle=\fontsize{7.5pt}{7.5pt}\ttfamily\selectfont,
  columns=fullflexible,
  breaklines=true,
  captionpos=b,
  commentstyle=\fontsize{7.5pt}{7.5pt}\color[rgb]{0.25,0.5,0.5},
  keywordstyle=\fontsize{7.5pt}{7.5pt}\color[rgb]{0.85,0.18,0.50},
}
\usepackage{etoolbox}
\makeatletter
\AfterEndEnvironment{algorithm}{\let\@algcomment\relax}
\AtEndEnvironment{algorithm}{\kern2pt\hrule\relax\vskip3pt\@algcomment}
\let\@algcomment\relax
\newcommand\algcomment[1]{\def\@algcomment{\footnotesize#1}}
\renewcommand\fs@ruled{\def\@fs@cfont{\bfseries}\let\@fs@capt\floatc@ruled
  \def\@fs@pre{\hrule height.8pt depth0pt \kern2pt}%
  \def\@fs@post{}%
  \def\@fs@mid{\kern2pt\hrule\kern2pt}%
  \let\@fs@iftopcapt\iftrue}
\makeatother

\usepackage[pagebackref,breaklinks,colorlinks]{hyperref}

\usepackage[capitalize]{cleveref}
\crefname{section}{Sec.}{Secs.}
\Crefname{section}{Section}{Sections}
\Crefname{table}{Table}{Tables}
\crefname{table}{Tab.}{Tabs.}

\def\cvprPaperID{0000} 
\def\confName{CVPR}
\def\confYear{2023}

\newcommand{\authsep}{\;\;}

\begin{document}

\title{Adaptive Texture-aware Masking for Self-Supervised Learning in 3D Dental CBCT Analysis}

\author{Xinquan Yang$_1$ \authsep 
	Jianfeng Ren$_2$ \authsep 
	Xuguang Li$_3$ \authsep 
	Kian Ming Lim$_2$ \authsep 
	He Meng$_3$ \\
Linlin Shen$^{\star}_1$ \authsep Yongqiang Deng$_3$\\
\\
xinquanyang99@gmail.com, llshen@szu.edu.cn}
\maketitle
{\let\thefootnote\relax\footnote{
{
$^{\quad\;\;\ 1}$ School of Artificial Intelligence, Shenzhen University, Shenzhen, China.
$^{\quad\;\;\ 2}$ School of Computer Science, University of Nottingham Ningbo China, Ningbo, China.\\
$^{\quad\;\;\ 3}$ Department of Stomatology, Shenzhen University General Hospital, Shenzhen, China.}}}

\vspace{-1em}
\begin{abstract}
Cone Beam Computed Tomography (CBCT) is pivotal for 3D diagnostic imaging in dentistry. However, the development of robust AI models for volumetric analysis is often constrained by the scarcity of large, annotated datasets. Self-supervised learning (SSL), particularly Masked Image Modeling (MIM), offers a promising pathway to leverage unlabeled data. A limitation of standard MIM is its reliance on random masking, which fails to prioritize diagnostically critical regions in dental CBCT volumes, such as subtle pathological changes and intricate anatomical boundaries.
To address this, we propose ATMask, a novel adaptive masking strategy. Instead of applying random masks or employing computationally intensive attention modules, ATMask computes an inter-slice texture variation map to identify regions with high structural or textural complexity. These high-variation areas are then selectively masked during pre-training, compelling the model to learn richer contextual representations essential for inferring complex 3D morphological transitions.
Furthermore, we contribute the first large-scale CBCT dataset, curated from both public and private sources, comprising 6,314 scans, for the dental AI model pretraining. Extensive experiments on three downstream dental CBCT tasks demonstrate that our ATMask enables more data-efficient and powerful representation learning than standard random masking and other advanced SSL baselines. 
The dataset and code will be released.
\end{abstract}

\section{Introduction}\label{sec:intro}
Cone Beam Computed Tomography (CBCT) has become an indispensable imaging modality in dentistry, providing detailed three-dimensional (3D) visualization of maxillofacial structures~\cite{yim2011analysis}. Unlike traditional two-dimensional techniques such as panoramic radiography and cephalometric X-rays, which project complex 3D anatomy onto a single plane and are susceptible to magnification and distortion~\cite{ketabi2025comparison}, CBCT offers high-resolution, isotropic 3D data. This allows for precise assessment of dental and craniofacial relationships, bone density, and anatomic variances. Its advantages like lower radiation dose compared to conventional CT (reported to be 50\%-90\% lower), relatively short scanning time, and capability for 1:1 true-size image reconstruction, have made CBCT the gold standard for complex diagnostic and treatment planning tasks~\cite{zaman2024comparing,mason2024systematic}. 

The proliferation of CBCT imaging has created unprecedented opportunities for developing artificial intelligence (AI) applications in dentistry. Consequently, AI models are being actively developed to assist in various tasks such as detecting periapical lesions and caries~\cite{jones2025dental, kwiatek2025comparison}, planning dental implants~\cite{ma2025preclinical,yang2024two,yang2023tceip}, and diagnosing maxillofacial pathologies~\cite{tassoker2025exploring}.
However, the development of robust, generalizable AI models for 3D CBCT volume analysis is hampered by a critical bottleneck: the scarcity of large-scale, meticulously annotated datasets. 
The process of manually segmenting anatomical structures or labeling pathologies slice-by-slice in a 3D CBCT volume is exceptionally time-consuming and requires specialized expertise~\cite{rao2024segmentation}. 
This high annotation cost results in limited in size and diversity for most publicly available CBCT datasets, which in turn can lead to overfitting and poor generalization of fully supervised models when faced with anatomical variations or imaging protocols not represented in the training data. 
To address this dependency on vast labeled datasets, researchers are turning to self-supervised learning (SSL) paradigms~\cite{pan2025structure,wu2024voco,xie2022simmim,he2022masked,he2020momentum}. These methods aim to learn meaningful representations directly from unlabeled data by solving a pretext task, thereby reducing the amount of annotated data required for effective fine-tuning on downstream tasks.

\begin{figure}[t]
	\centering
	\includegraphics[width=1.0\linewidth]{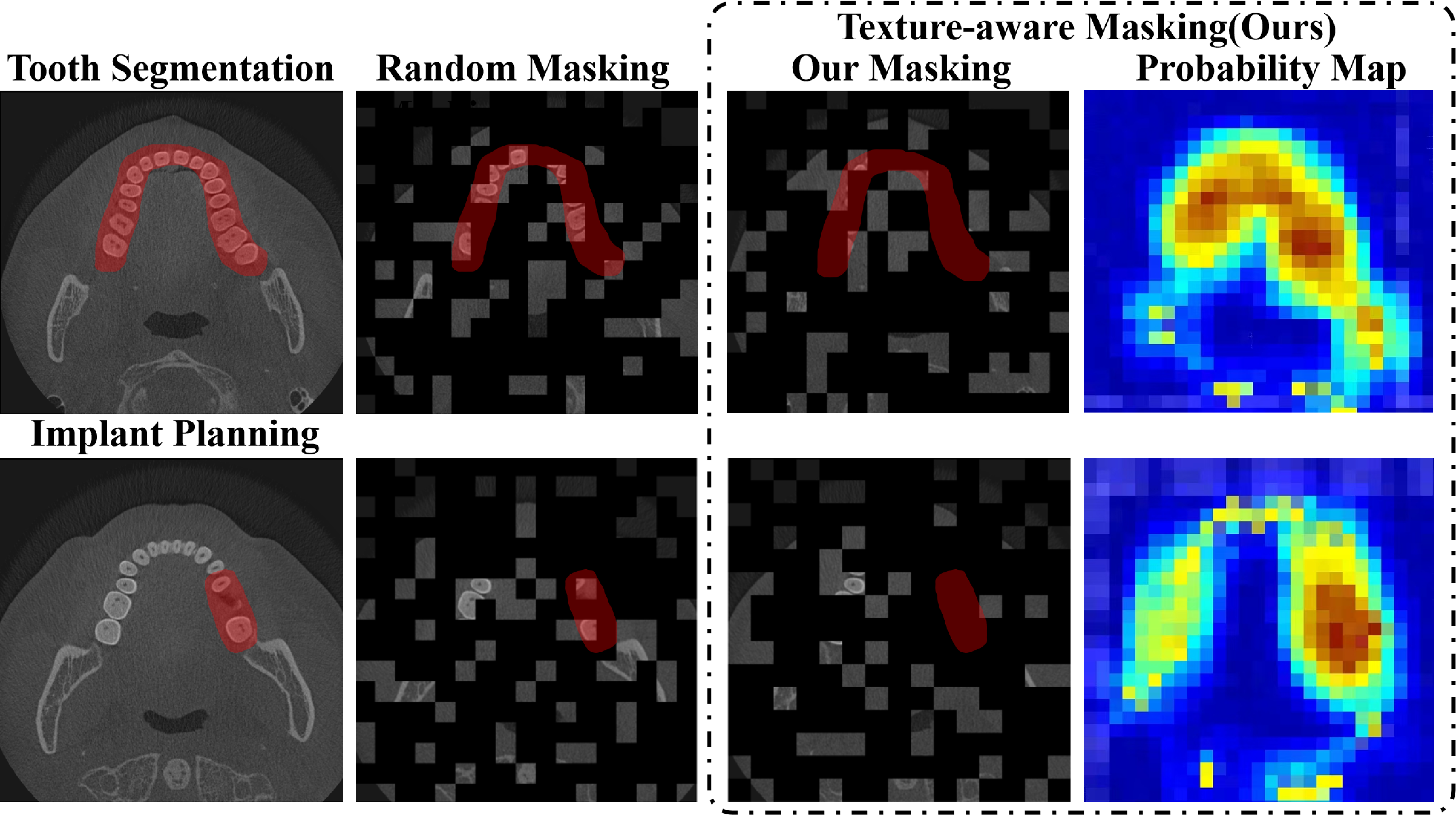}
	\caption{Comparison of our adaptive masking with existing random patch masking for masking ratio of 85\% (while black regions represent masked patches, gray ones are patches not masked). Our adaptive masking approach (third column) selects masks more patches within the task-related regions, to produce a domain-adapted challenging pretext task.}
	\label{fig_illustration}
\end{figure}

In 3D medical image analysis, prevalent SSL strategies can be broadly categorized into contrastive learning (CL)~\cite{xing2024hybrid,pan2025structure,chen2020simple,grill2020bootstrap} and masked image modeling (MIM)~\cite{chen2023masked,zhuang2025advancing,wang2023hard,gao2022mcmae}. 
CL aims to learn representations by maximizing agreement between differently augmented views of the same image instance while distinguishing them from other instances. 
While effective, this approach often requires careful selection of negative pairs. 
An alternative strategy is MIM, which involves randomly masking a portion of the input volume and training a model to reconstruct the missing content, forcing the model to learn robust anatomical and contextual priors.
Given the dense, contiguous nature of 3D anatomical structures in CBCT, MIM is a more natural and powerful fit for this modality than contrastive methods, as it directly learns the underlying generative structure of the data without relying on complex data augmentation or negative sampling. 
However, a significant limitation of standard MIM approaches is their reliance on random masking~\cite{xie2024rethinking,xie2023medim}. This uniform masking strategy treats all voxels equally, failing to prioritize regions of greater diagnostic significance, such as subtle pathological changes or intricate anatomical boundaries. 
Consequently, the model may spend considerable capacity learning to reconstruct diagnostically irrelevant homogeneous regions.
Although advanced methods~\cite{kakogeorgiou2022hide,xu2024self} have been proposed that employ attention mechanisms to prioritize the masking of informative regions, this approach often necessitates training an auxiliary network, which increases model complexity and can lead to training instability.

We observe that in many crucial dental applications—such as detecting periodontal disease, planning implant positions relative to critical anatomical structures, or segmenting thin pulp canals—the most diagnostically challenging and informative regions often exhibit significant textural variations across adjacent slices. 
These inter-slice variations are direct indicators of complex 3D morphological transitions within anatomical structures. 
This discovery encourages us to focus the mask region on areas with significant texture variations, thereby enabling the network to learn finer-grained features that are more relevant to the task (see Figure~\ref{fig_illustration}).
In this paper, we propose a novel adaptive masking strategy that directs the model's attention to these critical regions. Instead of computationally heavy attention mechanisms, our method computes a texture variation map across slices. 
Regions with high gradient and variance, indicative of significant structural or textural change between adjacent slices, are selectively masked. This forces the model to learn contextual features specifically tailored to inferring complex morphological transitions, which are critical for downstream dental AI tasks. 
By focusing the pre-training objective on these semantically meaningful regions, we aim to learn more powerful and data-efficient representations for 3D dental CBCT analysis, without the need for complex auxiliary networks.
The main contributions are summarized as follow:
\begin{itemize}
	\item We propose an adaptive masking strategy (ATMask) that enables the AI models to focus on the important region relative to the dental analysis tasks.
	\item We construct the first large-scale dental CBCT dataset comprising 6,314 scans to train the dental AI models, which greatly benefits the dental AI research community.
	\item Extensive experiments on three representative tasks, implant planning, tooth segmentation, and inferior alveolar nerve segmentation demonstrate the effectiveness of our proposed method.
\end{itemize}

\begin{figure*}[t]
	\centering
	\includegraphics[width=0.85\linewidth]{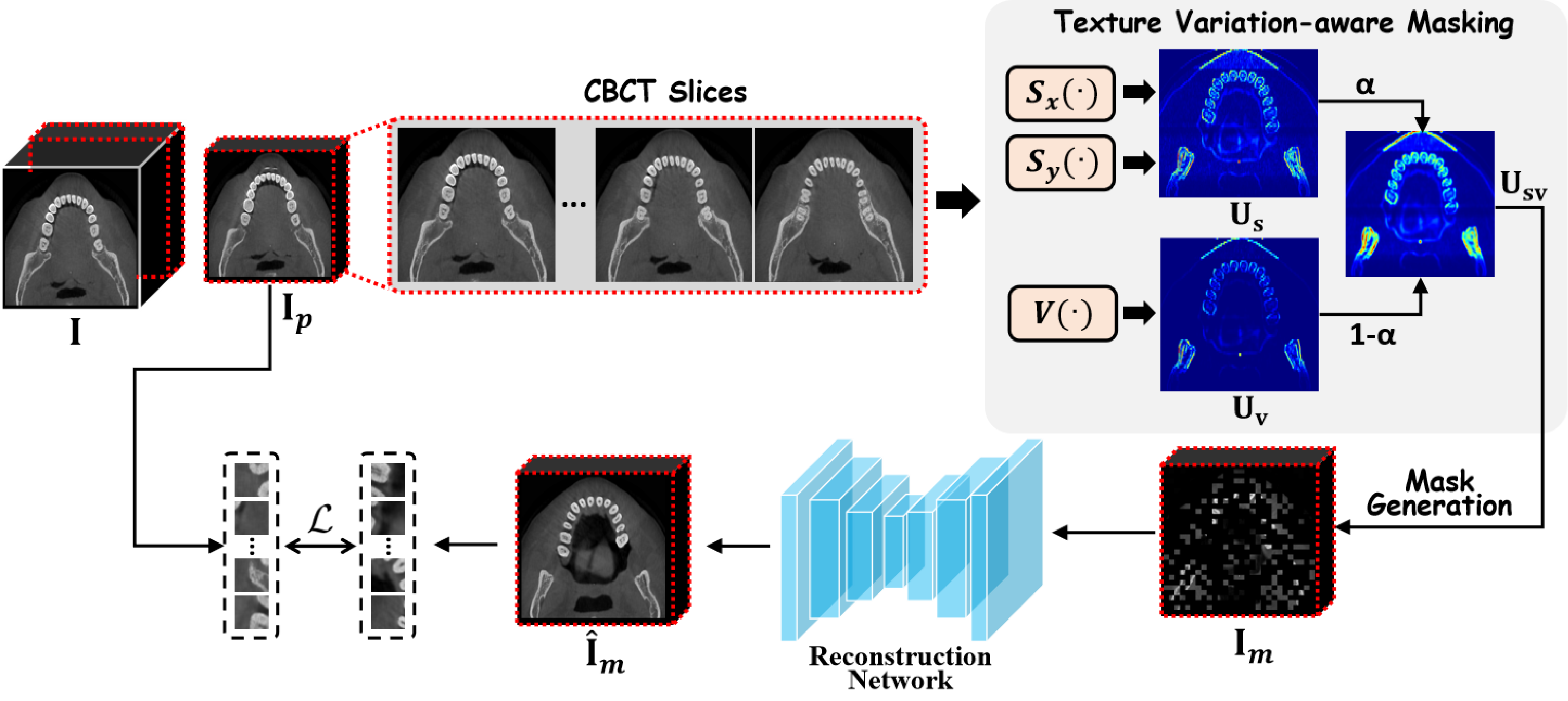}
	\caption{The architecture of the proposed ATMask.}
	\label{fig_network}
\end{figure*}


\section{Related work}\label{sec:relwork}
\subsection{Contrastive Learning Methods}
Contrastive learning has emerged as a dominant paradigm in self-supervised visual representation learning. Its core principle is to learn representations by contrasting positive pairs against negative pairs in a latent space. Many methods implement this by attracting different augmented views of the same image (positives) while repelling views from different images (negatives), as exemplified by SimCLR \cite{chen2020simple}. An alternative line of work employs clustering algorithms to generate consistent pseudo-labels for different views, thereby learning informative representations without explicit pairwise comparisons~\cite{caron2020unsupervised}. Furthermore, some recent advances have successfully removed the reliance on negative samples altogether. For instance, BYOL \cite{grill2020bootstrap} avoids collapse by predicting the output of one network from another using a mean squared error loss. While highly effective for learning global semantic features, a common limitation of these contrastive and relational methods is their relative weakness in capturing fine-grained, local visual patterns, which can be suboptimal for dense prediction tasks like segmentation.

\subsection{Masked Image Modeling Methods}
Masked Image Modeling (MIM) has recently gained prominence for its ability to learn rich, often fine-grained, representations by reconstructing masked portions of an input. These methods can be categorized based on the nature of the prediction target. One branch, initiated by BEiT \cite{bao2021beit}, predicts discrete visual tokens for masked patches. Another branch directly reconstructs continuous signal, such as raw pixels. Representative works include MAE \cite{he2022masked}, which employs an asymmetric encoder-decoder to predict a high proportion of randomly masked patches, and SimMIM \cite{xie2022simmim}, which simplifies the architecture for efficiency. Subsequent variants like ConvMAE \cite{gao2022mcmae} incorporate multi-scale convolutions to enrich the learned features. Other approaches explore more strategic masking. For example, AttMask \cite{kakogeorgiou2022hide} uses an attention map to guide the masking process. However, a key limitation of existing strategies is that the masking policy—whether random, fixed, or attention-based—is often not learned in an image-adaptive manner. This can lead to sub-optimal pre-training, particularly for specialized domains like medical imaging where informative regions are heterogeneous. The proposed ATMask addresses this by adaptive computing an inter-slice texture variation map to identify regions of high structural or textural variation.

\section{Methodology}
%
Figure~\ref{fig_network} illustrates the proposed architecture of our adaptive texture-aware masking network (ATMask), which consists of two main components: a texture variation-aware masking (TVM) module and a reconstruction network.
The TVM module operates in an unsupervised manner to identify regions with pronounced texture variations in CBCT images. By generating masks over these regions, the module encourages the reconstruction network to learn their distinctive features, which contributes to improved performance in downstream tasks.
Since the TVM module introduces no additional learnable parameters or architectural constraints at the network level, it is compatible with any feature extraction network. 
Next, we will elaborate on the details of these methods.

\subsection{Texture Variation-aware Masking}
Traditional MAE pre-training employs random masking, which treats all spatial regions equally during reconstruction.
However, for oral analysis tasks, the anatomical information and texture variations between 3D image slices are crucial for diagnosis, and a uniform random mask makes it difficult for the model to focus on these key areas.
Figure~\ref{fig_illustration} shows the masks generated by the random masking strategy on two oral analysis tasks: tooth segmentation and implant planning. The visualization clearly shows that the random masks often fail to cover anatomically critical regions (red region).	
To encourage the model to learn more robust representations by focusing on semantically challenging areas, we propose a texture-aware masking strategy. This method prioritizes the masking of image patches with high variation, which often correspond to tissue boundaries, textured regions, or areas with complex intensity variations.
The proposed strategy consists of two main stages: (1) calculating a 3D texture variation map, and (2) generating the final binary mask based on the texture variation distribution.

\subsubsection{Texture Variation Calculation.}
Given a 3D input volume $\mathbf{I} \in \mathbb{R}^{D \times H \times W}$, we compute a voxel-wise variation map $\mathbf{U} \in \mathbb{R}^{D \times H \times W}$ by aggregating multiple low-level image cues. To enhance continuity and computational efficiency, $\mathbf{U}$ is computed for groups of consecutive slices with a stride of $s$.
For a 2D slice $\mathbf{I}_z$ within a group, we compute its gradient magnitude map $\mathbf{G}_z$ and local variance map $\mathbf{V}_z$:
\begin{equation}
	\mathbf{G}_z = \sqrt{ (\mathbf{I}_z \ast \mathbf{S}_x)^2 + (\mathbf{I}_z \ast \mathbf{S}_y)^2 },
\end{equation}
\begin{equation}
	\mathbf{V}_z = \text{UniformFilter}(\mathbf{I}_z^2, w) - \left[\text{UniformFilter}(\mathbf{I}_z, w)\right]^2,
\end{equation}
where $\ast$ denotes convolution, $\mathbf{S}_x$ and $\mathbf{S}_y$ are Sobel operators, and $\text{UniformFilter}(\cdot, w)$ applies a mean filter with a window size of $w$. The slice-wise variation $\tilde{\mathbf{U}}_z$ is a weighted combination:
\begin{equation}
	\tilde{\mathbf{U}}_z = \alpha \cdot \hat{\mathbf{G}}_z + (1-\alpha) \cdot \hat{\mathbf{V}}_z.
\end{equation}
Here, $\hat{\mathbf{G}}_z$ and $\hat{\mathbf{V}}_z$ are normalized versions of $\mathbf{G}_z$ and $\mathbf{V}_z$ to the range $[0,1]$, and $\alpha$ is a weighting coefficient. 
The group map is obtained by taking the element-wise maximum across all slice variations within the group, which is then assigned to every slice in that group. After processing all groups, the resulting 3D map $\tilde{\mathbf{U}}$ is smoothed with a Gaussian filter (sigma=$\sigma$) and normalized to yield the final variation map $\mathbf{U}$.
The overall procedure is summarized in Algorithm \ref{alg_uncertainty}.

\begin{algorithm}
\caption{Compute 3D Texture Variation Map}
\label{alg_uncertainty}
\begin{algorithmic}[1]
\Require 3D CBCT scan $\mathbf{I} \in \mathbb{R}^{H \times W \times D}$, stride $s$, Gaussian blur parameter $\sigma$
\Ensure Texture variation map $\mathbf{U} \in \mathbb{R}^{H \times W \times D}$
		
\State Initialize variation map $\mathbf{U} \gets \mathbf{0}_{H \times W \times D}$
\State Get image depth $D \gets \text{height}(\mathbf{I})$
		
\For{$z = 0$ \textbf{to} $D-1$ \textbf{step} $s$}
  \If{$z + s \leq D$}
		\State Get slice group $\mathbf{G} \gets \mathbf{I}[z:z+s, :, :]$
		\State Initialize group map $\mathbf{G}_{\text{unc}} \gets \mathbf{0}_{W \times D}$
		
		\For{$i = 0$ \textbf{to} $|\mathbf{G}|-1$}
		\State Compute gradient: $\mathbf{G}_{\text{grad}} \gets \text{Gradient}(\mathbf{G}[i])$
		\State Compute variance: $\mathbf{G}_{\text{var}} \gets \text{Variance}(\mathbf{G}[i])$
		\State Compute slice variation: $\mathbf{S}_{\text{unc}} \gets \alpha \times \mathbf{G}_{\text{grad}} + (1-\alpha) \times \mathbf{G}_{\text{var}}$
		\State Update group map: $\mathbf{G}_{\text{unc}} \gets \max(\mathbf{G}_{\text{unc}}, \mathbf{S}_{\text{unc}})$
		\EndFor
		
		\For{$i = 0$ \textbf{to} $|\mathbf{G}|-1$}
		\State Apply group variation to entire group: $\mathbf{U}[z+i, :, :] \gets \mathbf{G}_{\text{unc}}$
		\EndFor
		\EndIf
		\EndFor
		
		\If{$\sigma > 0$}
		\State Apply Gaussian blur to variation map: $\mathbf{U} \gets \text{GaussianFilter}(\mathbf{U}, \sigma)$
		\EndIf
		
		\If{$\max(\mathbf{U}) > 0$}
		\State Normalize variation map: $\mathbf{U} \gets \mathbf{U} / \max(\mathbf{U})$
		\EndIf
		
		\State \Return $\mathbf{U}$
\end{algorithmic}
\end{algorithm}

\subsubsection{Texture-guided Mask Generation.}
Our goal is to generate a binary mask $\mathbf{M} \in \{0,1\}^{N_p}$ at the patch level, where $N_p$ is the total number of non-overlapping 3D patches. 
The overall masking ratio is $r$. We first downsample the texture variation map $\mathbf{U}$ to the patch grid resolution by average pooling within each patch, obtaining a patch-level texture score $u_i$ for the $i$-th patch.

Patches with $u_i > \tau$ are defined as high-variation regions, where $\tau$ is a pre-defined threshold. Let $N_h$ be the number of such high-variation patches. Our strategy allocates a portion $\beta$ of the total masks $m = \lfloor r \cdot N_p \rfloor$ to these regions. The number of masks assigned to high-variation patches is $m_h = \min(\lfloor \beta \cdot m \rfloor, N_h)$. The remaining masks $m_r = m - m_h$ are randomly selected from the rest of the patches.

The final binary mask $\mathbf{M}$ is constructed by first setting $m_h$ randomly chosen high-variation patches to 1 (masked), and then setting $m_r$ randomly chosen patches from the remaining pool to 1. This ensures a bias towards masking high-variation regions while maintaining some randomness for diversity. The patch-level mask is then upsampled to the full image resolution to obtain the final 3D mask applied to the input volume.

\subsection{Reconstruction Network}
The core of the masked autoencoder (MAE) based framework is the reconstruction network, which is tasked with predicting the original content of the masked regions based on the visible context. 
As our TAM module introduces no additional learnable parameters or architectural constraints at the network level, it is compatible with any reconstruction networks.

We define the reconstruction network as $\mathcal{F}_{\theta}$, parameterized by $\theta$.
It takes the masked input volume $\mathbf{I}_m = \mathbf{I} \odot (1 - \mathbf{M})$ as input, where $\odot$ denotes element-wise multiplication.
Its objective is to reconstruct the full target volume $\mathbf{\hat I}_m$:
\begin{equation}
	\mathbf{\hat I}_m = \mathcal{F}_{\theta}(\mathbf{I}_m)
\end{equation}

The network is trained to minimize the reconstruction error only on the masked patches. Given the predicted output $\mathbf{I}'$ and the target $\mathbf{T}$, the reconstruction loss $\mathcal{L}$ is computed as:
\begin{equation}
	\mathcal{L} = \frac{\sum_{i} \mathbf{M}_i \cdot \mathcal{L}_{\text{base}}(\mathbf{I}'_i, \mathbf{T}_i)}{\sum_{i} \mathbf{M}_i + \epsilon}
\end{equation}
where $\mathcal{L}_{\text{base}}$ is the per-voxel base loss function (Mean Squared Error), $i$ indexes all spatial locations, and $\epsilon$ is a small constant for numerical stability. 
This masking-and-reconstruction pre-training task forces the model to develop a comprehensive understanding of the underlying anatomical patterns and their contextual relationships within 3D medical volumes.

\begin{table*}[htbp]
	\centering
	\caption{Summary of CBCT-based Studies and Available Data}
	\label{tab_dataset}
	\resizebox{0.8\linewidth}{!}{ 
		\begin{tabular}{lllllr}
			\toprule
			\rowcolor{gray!20} 
			\textbf{Modality} & \textbf{Authors} & \textbf{Name} & \textbf{Year} & \textbf{Country} & \textbf{Available Data} \\
			\midrule
			\multirow{7}{*}{CBCT} & Cipriano et al.~\cite{cipriano2022deep} & -- & 2022 & Italy & 347 \\
			& Cui et al.~\cite{cui2022ctooth} & CTooth & 2022 & China & 37 \\
			& Cui et al.~\cite{cui2022fully} & -- & 2022 & China & 150 \\
			& Bolelli et al.~\cite{bolelli2024segmenting} & ToothFairy & 2023 & Italy, Netherlands & 493 \\
			& Bolelli et al.~\cite{bolelli2025segmenting} & ToothFairy2 & 2024 & Italy, Netherlands & 530 \\
			& Wang et al.~\cite{wang2025miccai} & STSR & 2025 & China & 730 \\
			& Ours & Private Data & 2026 & China & 3405 \\
			\midrule
			\rowcolor{gray!10} 
			\multicolumn{3}{l}{\textbf{Data released in this paper}} & & Italy, Netherlands, China & \textbf{6314} \\
			\bottomrule
	\end{tabular}}
\end{table*}

\subsection{Pretraining Dataset}
To facilitate the development of the dental AI community, we have collected the first large-scale CBCT dataset.
The dataset is a composite of multiple publicly available datasets and a substantial private collection, ensuring both breadth and depth in the representation of dental and craniofacial anatomies. A comprehensive summary of the data sources is provided in Table~\ref{tab_dataset}.

\textbf{Public Datasets.} 
We incorporate several well-established public CBCT datasets to enhance generalizability. These include: 1) 347 scans from the work of~\cite{cipriano2022deep}; 2) The CTooth dataset~\cite{cui2022ctooth} (37 scans) and an additional 150 scans from~\cite{cui2022fully}; 3) The ToothFairy (493 scans) and its extension ToothFairy2 (530 scans) datasets from \cite{bolelli2024segmenting, bolelli2025segmenting}; and 4) The STSR dataset (730 scans) from \cite{wang2025miccai}. Collectively, these public sources contribute 2,287 CBCT volumes from studies conducted across different countries (Italy, China, Netherlands) between 2021 and 2025, covering a variety of clinical presentations and scanning protocols.

\textbf{Private Dataset.} 
To increase the data scale and diversity, we further collected a large private dataset, which consists of 3,405 anonymized CBCT scans. The private dataset encompasses a wide range of clinical indications, patient demographics, and scanner models, which is crucial for training a robust model that can handle real-world variability.

\textbf{Data Preprocessing.} 
All CBCT volumes underwent a standardized preprocessing pipeline. First, voxel intensities were clipped to the Hounsfield Unit (HU) window of $[-1000, 2000]$ to focus on the relevant tissue range. 
Subsequently, intensity normalization was applied to each volume to have zero mean and unit variance. To manage computational load and conform to the network input size, all volumes were resampled to an isotropic resolution of 0.5 mm. 

\begin{table}[htbp]
\centering
\begin{minipage}{0.48\textwidth}
		\centering
		\caption{Experimental results on the task of tooth segmentation. $\textsuperscript{$\dag$}$ denotes we re-implement the approach.}
		\label{tab:tooth_segmentation}
		\resizebox{\linewidth}{!}{
			\begin{tabular}{lccccc}
				\toprule
				\rowcolor{gray!20}
				\textbf{Model} & \textbf{Network} & \textbf{Dice(\%)} & \textbf{IoU} & \textbf{HD95} \\
				\midrule
				\rowcolor{gray!20}
				\multicolumn{5}{l}{From Scratch} \\
				\midrule
				- & UNet     & 73.5 & 0.642 & 4.14 \\
				- & UNet++   & 74.6 & 0.663 & 4.26 \\
				- & UXNet    & 71.9 & 0.619 & 4.57 \\
				- & UNETR    & 64.9 & 0.526 & 5.45 \\
				- & UNETR++  & 72.1 & 0.618 & 3.93 \\
				- & SwinUNETR & 72.5 & 0.634 & 4.05 \\
				\midrule
				\rowcolor{gray!20}
				\multicolumn{5}{l}{With CBCT SSL} \\
				\midrule
				\multirow{3}{*}{MAE~\cite{he2022masked}} & UNet      & 74.3 & 0.652 & 4.27 \\
				& UNETR     & 65.8 & 0.538 & 5.58 \\
				& SwinUNETR & 73.1 & 0.636 & 4.41 \\
				\midrule
				PCRLv2~\cite{zhou2023unified}      & SwinUNETR & 72.9 & 0.640 & 4.18 \\
				SwinMM~\cite{wang2023swinmm}      & SwinUNETR & 73.8 & 0.649 & 4.19 \\
				AttMask\textsuperscript{$\dag$}~\cite{kakogeorgiou2022hide}     & SwinUNETR & 72.9 & 0.637 & 4.53 \\
				$S^2DC$\textsuperscript{$\dag$}~\cite{pan2025structure}      & SwinUNETR & 73.2 & 0.643 & 4.24 \\
				VoCo~\cite{wu2024voco}        & SwinUNETR & 73.7 & 0.644 & 4.29 \\
				\midrule
				\multirow{3}{*}{ATMask(Ours)} & UNet      & \textbf{74.7} & \textbf{0.657} & 3.63 \\
				& UNETR     & 66.2 & 0.542 & 5.31 \\
				& SwinUNETR & 74.1 & 0.651 & \textbf{3.36} \\
				\bottomrule
			\end{tabular}
		}
	\end{minipage}
	\hfill
	\begin{minipage}{0.48\textwidth}
		\centering
		\caption{Experimental results on the task of IAN segmentation. $\textsuperscript{$\dag$}$ denotes we re-implement the approach.}
		\label{tab:IAN_segmentation}
		\resizebox{\linewidth}{!}{
			\begin{tabular}{lccccc}
				\toprule
				\rowcolor{gray!20}
				\textbf{Model} & \textbf{Network} & \textbf{Dice(\%)} & \textbf{IoU} & \textbf{HD95} \\
				\midrule
				\rowcolor{gray!20}
				\multicolumn{5}{l}{From Scratch} \\
				\midrule
				- & UNet & 76.88 & 0.627 & 13.93 \\
				- & UNet++ & 77.53 & 0.636 & 13.34 \\
				- & UXNet & 72.67 & 0.575 & 16.74 \\
				- & UNETR & 39.63 & 0.252 & 93.28 \\
				- & UNETR++ & 77.92 & 0.641 & 13.98 \\
				- & SwinUNETR & 76.35 & 0.621 & 13.34 \\
				\midrule
				\rowcolor{gray!20}
				\multicolumn{5}{l}{With CBCT SSL} \\
				\midrule
				\multirow{2}{*}{MAE~\cite{he2022masked}} & UNet & 77.92 & 0.641 & 13.98 \\
				& SwinUNETR & 76.93 & 0.628 & 13.35 \\
				\midrule
				PCRLv2~\cite{zhou2023unified} & SwinUNETR & 77.29 & 0.633 & 14.77 \\
				SwinMM~\cite{wang2023swinmm} & SwinUNETR & 77.13 & 0.631 & 13.59 \\
				AttMask\textsuperscript{$\dag$}~\cite{kakogeorgiou2022hide} & SwinUNETR & 76.67 & 0.625 & 14.09 \\
				$S^2DC$\textsuperscript{$\dag$}~\cite{pan2025structure} & SwinUNETR & 76.93 & 0.628 & 13.35 \\
				VoCo~\cite{wu2024voco} & SwinUNETR & 77.55 & 0.636 & 12.88 \\
				\midrule
				\multirow{2}{*}{ATMask(Ours)} & UNet & 78.57 & 0.651 & \textbf{11.83} \\
				& SwinUNETR & \textbf{78.87} & \textbf{0.654} & 11.98 \\
				\bottomrule
			\end{tabular}
		}
	\end{minipage}
\end{table}

\section{Experiments}
\subsection{Downstream Datasets}
We selected three commonly used CBCT oral analysis tasks to form a comprehensive evaluation suite that validates the effectiveness of our ATMask from complementary perspectives. 
The three tasks are tooth segmentation, inferior alveolar nerve segmentation, and dental implant planning, all of which use publicly available datasets.

\textbf{Tooth Segmentation.} 
This task serves as a primary test for the model's ability to capture fine-grained local textures and precise boundaries. 
High-variation regions in CBCT scans frequently coincide with intricate interfaces such as enamel-dentine junctions, interproximal contacts, and the subtle boundary between tooth and alveolar bone. 
A pre-training strategy that forces the model to reconstruct these complex textures should yield features that significantly improve the accuracy of delineating individual tooth crowns and roots, especially at their boundaries.
We used the dataset proposed by Cui \etal~\cite{cui2022fully}, 
which contains 150 CBCT scans, for evaluation. 
We use 80\% for training and 20\% for testing.

\textbf{Inferior Alveolar Nerve Segmentation.} 
This task evaluates the model's capability to learn from low-contrast and structurally ambiguous regions. The mandibular canal is a critical yet often poorly contrasted structure in CBCT. Our variation map, heavily influenced by gradient and variance, highlights these faint, tubular regions. 
By concentrating the reconstruction effort there during pre-training, the model reduces false negatives in ambiguous areas.
We used the ToothFairy2~\cite{bolelli2025segmenting} dataset, which contains 145 finely annotated CBCT scans.
We use 80\% for training and 20\% for testing.

\textbf{Dental Implant Planning.} 
This high-level planning task assesses the model's integrated understanding of global anatomy and spatial relationships. Successful planning requires simultaneously reasoning about the bone density and morphology at the implant site, the position of adjacent teeth, and the precise 3D path of the mandibular canal to avoid injury. Our masking strategy, by focusing on regions with complex local information (bone texture, canal edges) and their context, encourages the learning of holistic features that encode the spatial interplay between different anatomical entities. 
We use the ImplantFairy~\cite{yang2026regfreenet} dataset, which contains 1622 finely annotated CBCT scans. We followed the official dataset split, which contains 1369 training data and 253 test data.

In summary, tooth segmentation validates the gain in local discriminative power for fine details; inferior alveolar nerve segmentation probes the improvement in handling low-signal, high-ambiguity structures; and implant planning tests the enhancement in global anatomical reasoning. The consistent improvement across these diverse and challenging tasks would provide strong evidence that our ATMask pre-training paradigm successfully guides the model to learn more transferable and semantically meaningful representations from unlabeled CBCT data.

\subsection{Training setup}
Our experiments are implemented in PyTorch and utilize the MONAI framework. The training is conducted in a distributed manner using 8 NVIDIA H100 GPUs. 
The pipeline consists of two main stages: (1) self-supervised pre-training of a Masked Autoencoder (MAE) and (2) fully supervised fine-tuning for a downstream segmentation task. 
During pre-training stage, the model is trained for 30 epochs with a per-GPU batch size of 5. The AdamW optimizer is used with a learning rate of $1.5\times10^{-4}$ and a weight decay of 0.05. A linear warmup for 10 epochs followed by a cosine annealing schedule is applied. 
The input 3D volumes are processed with a patch size of 16 and normalized to a range of [0, 1] after intensity scaling from $[-1000, 500]$ Hounsfield Units (HU). 
The reconstruction loss is the Mean Squared Error (MSE) loss, applied only on the masked patches.
In fine-tuning stage, the model is trained for 50 epochs. The AdamW optimizer is used with a base learning rate of $1\times10^{-4}$, which is linearly scaled according to the total effective batch size. 
A weight decay of $1\times10^{-5}$ is applied. The learning rate scheduler uses a warmup for 5 epochs followed by cosine annealing. The per-GPU batch size is 1, and the Dice-Cross-Entropy (DiceCELoss) loss function is used for optimization. 
Input volumes are spatially normalized to an isotropic spacing of 1mm and cropped/padded to a fixed size of $128 \times 128 \times 128$. Standard data augmentations including random spatial cropping and padding are applied during training. The sliding window inference during validation uses a batch size of 4 and an overlap of 0.5.

\subsection{Evaluation Criteria}
To validate the effectiveness of our method, we use the Dice Similarity Coefficient (DSC), Intersection over Union (IoU) and 95\% Hausdorff Distance (HD95). 
Note that higher values for these metrics, except HD95.
The detail definitions are given in the \textbf{Supplementary material}.
Formally,
\begin{align}
	\mathrm{DSC} &= \frac{2 \times \mathrm{TP} + \epsilon}{T + P + \epsilon}, \\[6pt]
	\mathrm{IoU} &= \frac{\mathrm{TP} + \epsilon}{T + P - \mathrm{TP} + \epsilon}, \\[6pt]
	\mathrm{HD95} &= \max_{95\%}\bigl(d(P,G),\, d(G,P)\bigr)
\end{align}


where $\epsilon$ is a small constant to avoid zero division. TP, FP, and FN are the number of true positive points, false positive points, and false negative points, respectively. 
$T$ is the number of ground-truth points of that class, $P$ is the number of predicted positive points, $G$ is the number of ground-truth positive points. 

\subsection{Experiments on Downstream Tasks}
\subsubsection{Tooth Segmentation.}
In Table~\ref{tab:tooth_segmentation}, we compare the performance of different pretrained methods in the task of tooth segmentation. 
We evaluate multiple 3D segmentation architectures under two training regimes: training from scratch and fine-tuning with CBCT-based self-supervised learning (SSL) pre-training. 
In the from-scratch setting, UNet++ achieves the highest Dice score of 74.6\% and IoU of 0.663, while UNETR++ yields the best boundary accuracy with an HD95 of 3.93. 
When models are initialized with CBCT SSL pre-training, consistent improvements are observed across most architectures. Notably, our proposed ATMask method, when combined with the SwinUNETR backbone, achieves the overall best performance with a Dice score of 74.1\%, IoU of 0.651, and a notably low HD95 of 3.36. This represents a significant gain over the from-scratch SwinUNETR (Dice: 72.5\%, IoU: 0.634, HD95: 4.05) and also surpasses other contemporary SSL methods like SwinMM (Dice: 73.8\%) and VoCo (Dice: 73.7\%) when using the same backbone. 

\subsubsection{Inferior Alveolar Nerve Segmentation.}
The experimental results for the Inferior Alveolar Nerve (IAN) segmentation task are detailed in Table~\ref{tab:IAN_segmentation}. When trained from scratch, the UNETR++ model achieved the best performance among all compared architectures, with a Dice score of 77.92\%, an IoU of 0.641, and an HD95 of 13.98. The introduction of CBCT-based self-supervised learning (SSL) pre-training leads to a general performance improvement across most models. Notably, our proposed ATMask framework demonstrates superior capability in leveraging such pre-training. When integrated with the UNet backbone, ATMask attains a Dice score of 78.57\%, an IoU of 0.651, and a significantly reduced HD95 of 11.83. With the SwinUNETR backbone, it further achieves the highest Dice score of 78.87\% and IoU of 0.654, while maintaining a low HD95 of 11.98. These results not only surpass all from-scratch baselines but also outperform other contemporary SSL methods under the same backbone (e.g., VoCo with Dice: 77.55\%, HD95: 12.88). The substantial reduction in HD95, in particular, underscores ATMask's effectiveness in producing more precise boundary delineation for the challenging IAN structure, which is critical for clinical safety in dental procedures.

\subsubsection{Dental Implant Planning.}
We validate the effectiveness of ATMask in the task of dental implant planning in Table~\ref{tab:implant_planning}.
The experimental results demonstrate the significant advantage of the proposed self-supervised pre-training method for the implant position prediction task. While models trained from scratch show limited performance, with CNN-based UNet (76.88\% Dice) outperforming most Transformers, our method consistently achieves state-of-the-art results across different backbones after pre-training. It elevates UNet to 78.57\% Dice and SwinUNETR to 78.87\% Dice, surpassing all compared SSL baselines (e.g., VoCo: 77.55\%). This indicates that our approach effectively leverages unlabeled CBCT data to learn superior, transferable representations that are specifically beneficial for this medical imaging task, offering a robust solution to data scarcity.

\begin{table}[htbp]
	\centering
	\begin{minipage}{0.5\textwidth}
		\centering
		\caption{Experimental results on the task of dental implant planning. $\textsuperscript{$\dag$}$ denotes we re-implement the approach.}
		\label{tab:implant_planning}
		\resizebox{0.85\linewidth}{!}{
			\begin{tabular}{lccc}
				\toprule
				\rowcolor{gray!20}
				\textbf{Model} & \textbf{Network} & \textbf{Dice(\%)} & \textbf{IoU} \\
				\midrule
				\rowcolor{gray!20}
				\multicolumn{4}{l}{From Scratch} \\
				\midrule
				- & UNet & 45.48 & 0.336 \\
				- & UNet++ & 43.75 & 0.316 \\
				- & UXNet & 42.09 & 0.309 \\
				- & UNETR & 41.34 & 0.282 \\
				- & UNETR++ & 43.80 & 0.316 \\
				- & SwinUNETR & 44.03 & 0.321 \\
				\midrule
				\rowcolor{gray!20}
				\multicolumn{4}{l}{With CBCT SSL} \\
				\midrule
				\multirow{3}{*}{MAE~\cite{he2022masked}} & UNet & 46.33  & 0.336 \\
				& UNETR & 43.48 & 0.315 \\
				& SwinUNETR & 45.36 &0.334 \\
				\midrule
				PCRLv2~\cite{zhou2023unified} & SwinUNETR &46.39 & 0.339 \\
				SwinMM~\cite{wang2023swinmm} & SwinUNETR & 46.82 & 0.341\\
				AttMask\textsuperscript{$\dag$}~\cite{kakogeorgiou2022hide} & SwinUNETR & 46.17 & 0.337 \\
				$S^2DC$\textsuperscript{$\dag$}~\cite{pan2025structure} & SwinUNETR & 46.75 & 0.340\\
				VoCo~\cite{wu2024voco} & SwinUNETR & 47.10 & 0.345\\
				\midrule
				\multirow{3}{*}{ATMask(Ours)} & UNet & \textbf{47.68} & \textbf{0.352}\\
				& UNETR & 44.83 & 0.325 \\
				& SwinUNETR & 47.27 & 0.344 \\
				\bottomrule
		\end{tabular}}
	\end{minipage}
	\hfill
	\begin{minipage}{0.48\textwidth}
		\centering
		\includegraphics[width=\linewidth]{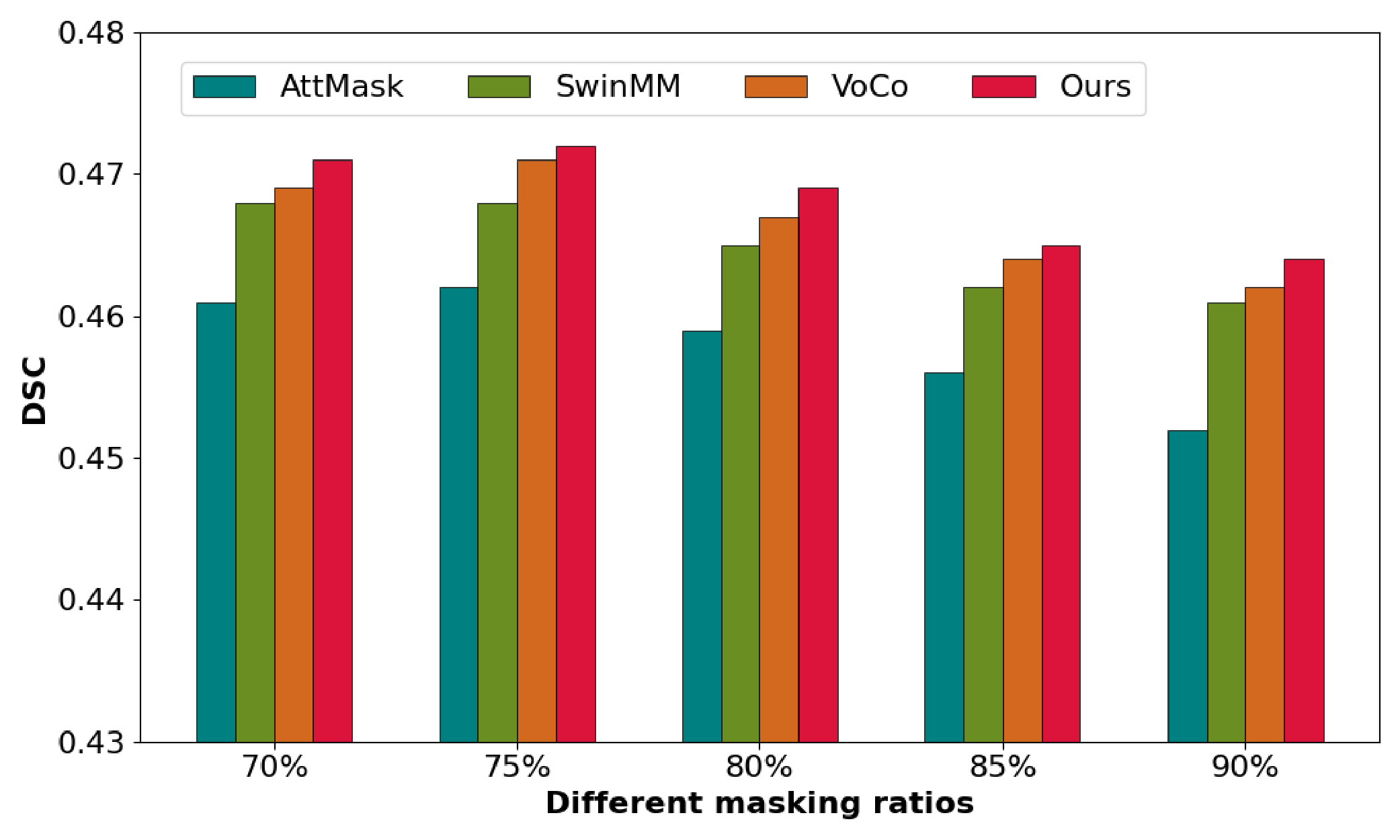}
		\captionof{figure}{Comparison of different masking ratios for various self-supervision methods.}
		\label{fig_mr}
		
		\captionof{table}{Ablation study of different masking ratios in high-variation regions.}
		\centering
		\resizebox{0.8\linewidth}{!}{
			\begin{tabular}{ccccc}
				\toprule
				\rowcolor{gray!20}
				\textbf{Method} & \textbf{Mask Ratio} & \textbf{DSC (\%)} & \textbf{IoU} \\
				\midrule
				\multirow{6}{*}{ATMask}
				& 0.60  & 47.22 & 0.341 \\ 
				& 0.65  & \textbf{47.68} & \textbf{0.352} \\
				& 0.70  & 47.01 & 0.348 \\
				& 0.75  & 46.51 & 0.342 \\
				& 0.80  & 45.85 & 0.335 \\
				& 0.85  & 45.46 & 0.335 \\
				\bottomrule
		\end{tabular}}
		
		\label{tab:mask_ratio_ablation}
	\end{minipage}
\end{table}

\subsection{Different Masking Ratios}
We further investigate the impact of high masking ratios on segmentation performance by comparing our method with AttMask, SwinMM, and VoCo across ratios from 70\% to 90\%. The results are depicted in Figure~\ref{fig_mr}. A key observation is that all methods, including ours, achieve their peak performance at a masking ratio of 75\%, with our method securing the highest DSC of approximately 0.47. This suggests that 75\% represents an optimal balance for the masking-based pre-training paradigm, providing sufficient contextual challenge for the model to learn robust representations without overly compromising the integrity of the input information. Beyond this point, the performance of all methods declines as the masking ratio increases to 85\% and 90\%. This decline can be attributed to the excessive removal of visual tokens, which likely hinders the model's ability to reconstruct meaningful and discriminative features during pre-training, thereby limiting the transferable knowledge for the downstream segmentation task. 
Notably, our method consistently outperforms the competitors across all tested high ratios. This superior and more stable performance underlines the effectiveness of our masking strategy, which is adept at identifying and preserving critical information within the limited unmasked regions (e.g., 25\% at 75\% ratio), even under extremely high masking conditions.

\subsection{Effects on Masking Ratios in High Variation Regions}
Determining the optimal masking ratio for high-variation regions is a crucial aspect of our ATMask framework. To this end, we conducted an ablation study on the ImplantFairy dataset. As shown in Table~\ref{tab:mask_ratio_ablation}, the results reveal a clear performance trade-off. Our model achieves its peak performance (DSC: 47.68\%, IoU: 0.352) at a masking ratio of 0.65. This suggests that masking approximately 65\% of the most variation patches constitutes an ideal pretext task for self-supervised pre-training; it provides sufficient challenge for the reconstruction network to learn robust features from these complex regions, without being impaired by excessive information loss. Beyond this optimal point, further increasing the masking ratio leads to a consistent decline in both DSC and IoU. This degradation indicates that over-masking eventually removes too much essential structural and contextual information from these critical areas. In summary, this ablation study confirms that the mechanism for identifying high-variation regions in ATMask is most effective when combined with a moderate masking intensity.

\begin{figure*}[t]
	\centering
	\includegraphics[width=0.7\linewidth]{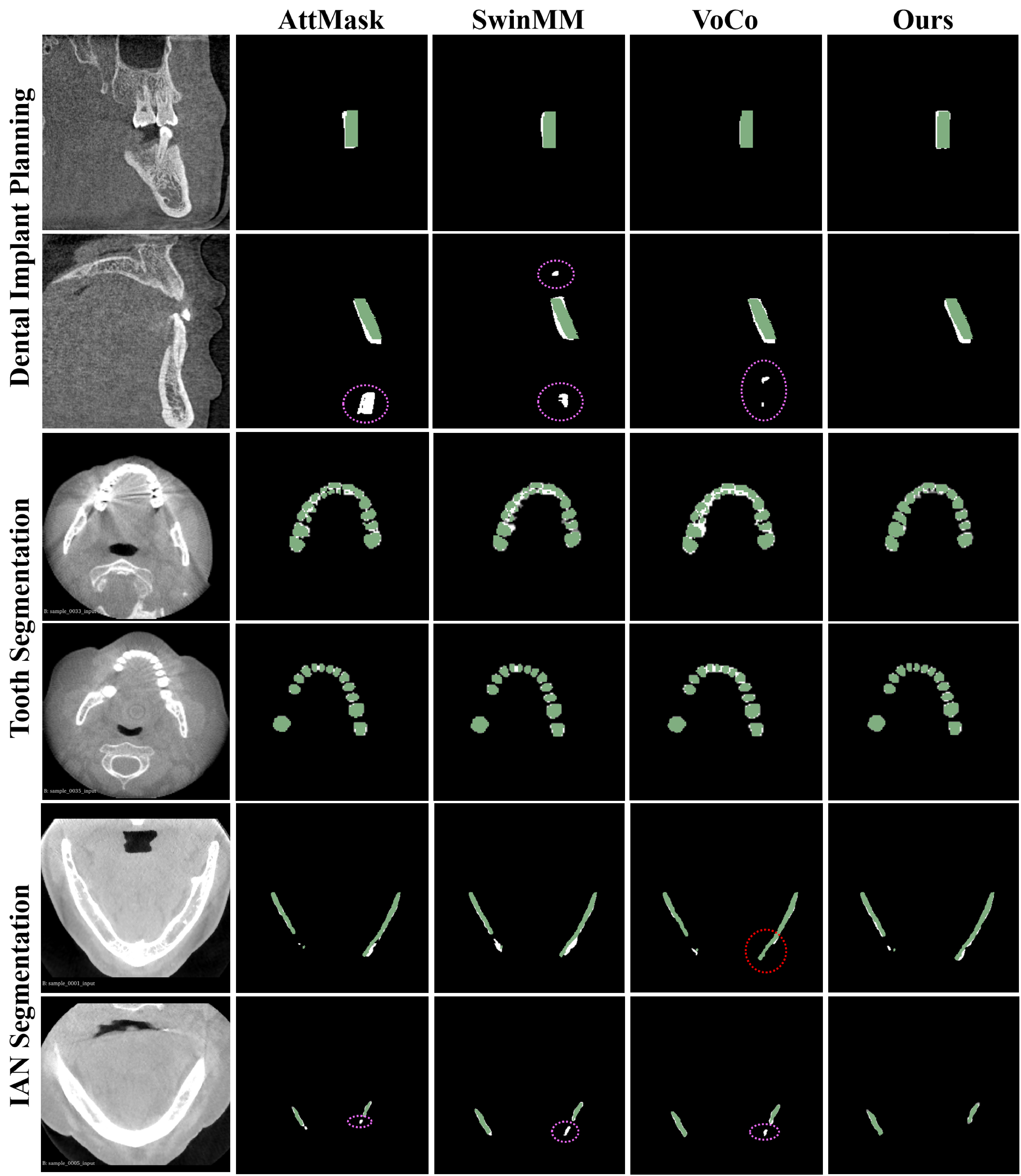}
	\caption{Visual comparison of different methods under various downstream tasks.}
	\label{fig_vis}
\end{figure*}

\subsection{Sensitivity Analysis}
There are two hyperparameters ($\alpha$ and $\beta$) in the calculation of the texture variation map. Sensitivity analyses of these two parameters are performed on the ImplantFairy dataset, and the evaluation results are presented in Fig.1. 
The effects of the two hyperparameters differ significantly. 
The DSC value exhibits a unimodal trend with respect to A, increasing to a peak at $\alpha$=0.6 before decreasing as $\alpha$ rises further to 0.9. Conversely, the relationship with $\beta$ is monotonically negative, with DSC consistently declining as $\beta$ increases from 0.1 to 0.5.

\begin{figure*}
	\centering
	\includegraphics[width=0.9\linewidth]{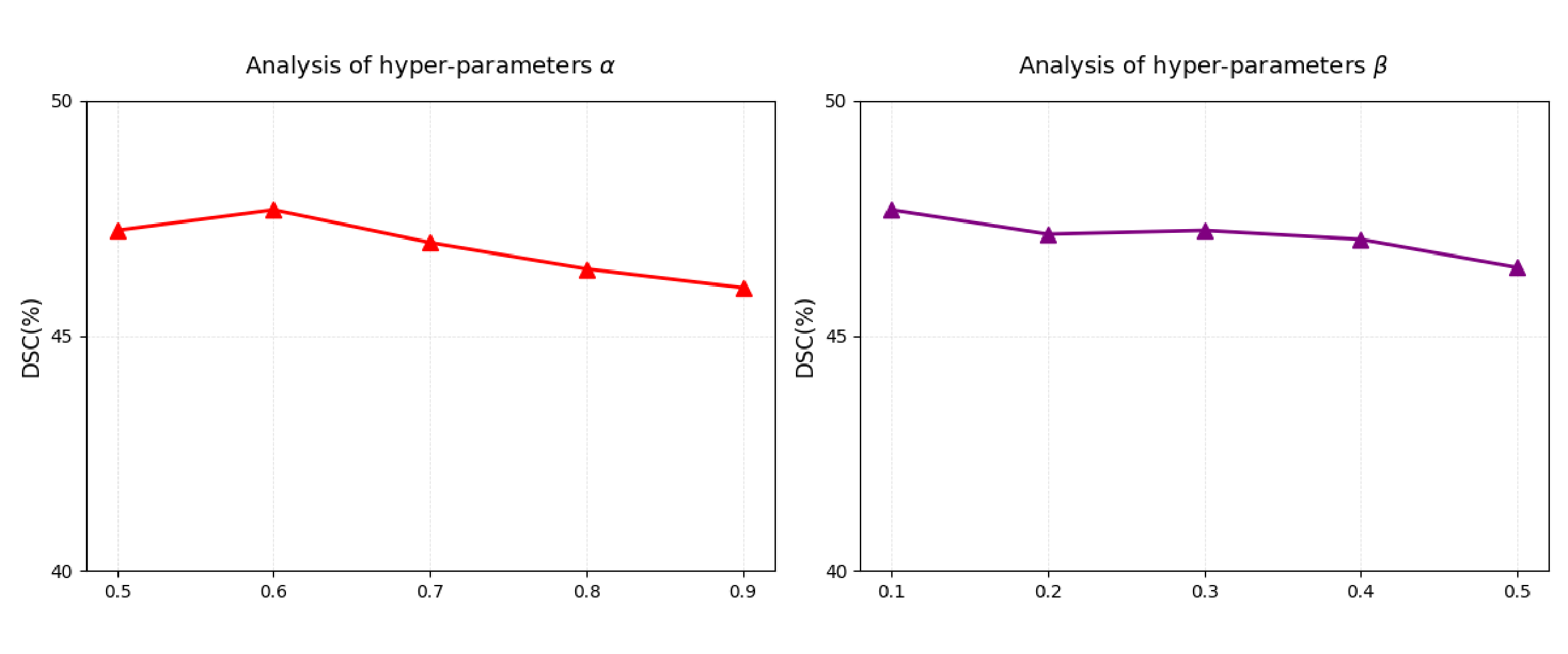}
	\caption{Sensitivity analyses of hyper-parameters $\alpha$ and $\beta$. The Dice Similarity Coefficient (DSC) values here are reported on the ImplantFairy dataset.}
	\label{fig_sensitivity}
\end{figure*}

\subsection{Visualization Analysis}
Figure~\ref{fig_vis} presents qualitative results comparing the proposed method with three existing approaches: AttMask, SwinMM, and VoCo, for three tasks: dental implant planning, tooth segmentation, and inferior alveolar nerve (IAN) segmentation. For each example, the left sub-figure shows the original medical scan, and the right sub-figure overlays the model prediction (in white) with the ground truth (in green).

For the dental implant planning task, the first case (top row) involves a standard implant placement. Here, the proposed method demonstrates more precise trajectory localization and better anatomical alignment. In contrast, AttMask and VoCo exhibit noticeable orientation deviations, while SwinMM produces coarser boundaries. The second case (second row) presents a more complex scenario with a tilted implant. The results show that the proposed method accurately predicts both the position and orientation of the implant. The other methods, however, generate incorrect predictions (purple circles), with VoCo performing particularly poorly in estimating the implant inclination.
For the tooth segmentation task, two challenging cases with metal artifacts were selected. The proposed approach effectively captures fine morphological details and maintains boundary continuity, with its predictions closely matching to the ground truth. AttMask yields fragmented predictions in certain regions, SwinMM tends to over-segment, and VoCo under-segments subtle structures such as dental crowns.
In the Inferior Alveolar Nerve (IAN) segmentation task, which requires precise delineation of a thin and tortuous nerve canal, our method achieves the highest overlap with the ground truth, particularly in challenging low-contrast regions. 
In contrast, AttMask and SwinMM produce discontinuous segmentations along the nerve trajectory (indicated by purple circles), while VoCo exhibits a clear case of false connection or over-segmentation, erroneously bridging a canal branch (marked by a red circle).

Overall, the visual comparisons confirm that the proposed method delivers superior performance in capturing anatomical details, maintaining structural consistency, and aligning accurately with expert annotations across all three clinical tasks. The improvements are most evident in challenging areas characterized by thin structures, low-contrast boundaries, and complex morphological variations.

\section{Conclusion}
In this paper, we propose a novel adaptive masking strategy (ATMask) for self-supervised learning in 3D dental CBCT analysis, which computes an inter-slice texture variation map to identify regions of high structural or textural complexity and selectively masking them during pre-training. 
This process forces the model to learn richer contextual representations of complex 3D morphological transitions. 
Further, we contribute a large-scale CBCT dataset comprising 6,314 CBCT scans. Extensive experiments on three downstream tasks confirm that our ATMask yields more data-efficient and powerful representations compared to standard random masking and other advanced SSL baselines.

\bibliographystyle{ieee_fullname}
\bibliography{egbib}

\end{document}